\documentclass[10pt,twocolumn,letterpaper]{article}

\usepackage{wacv}
\usepackage{times}
\usepackage{epsfig}
\usepackage{graphicx}
\usepackage{amsmath}
\usepackage{amssymb}

\def\wacvPaperID{656} % Enter the WACV Paper ID here

\wacvfinalcopy % *** Uncomment this line for the final submission

\ifwacvfinal
\def\assignedStartPage{9876} % *** Enter the assigned starting page number (instead of 9876)
\fi

\ifwacvfinal
\usepackage[breaklinks=true,bookmarks=false]{hyperref}
\else
\usepackage[pagebackref=true,breaklinks=true,colorlinks,bookmarks=false]{hyperref}
\fi

\ifwacvfinal
\else
\fi

\begin{document}

%%%%%%%%% TITLE
\title{Understanding Egocentric Hand-Object Interactions from Hand Pose Estimation}

\author{Yao LU\\
Bristol University\\
% Institution1 address\\
{\tt\small yl1220@brsitol.ac.uk}
% For a paper whose authors are all at the same institution,
% omit the following lines up until the closing ``}''.
% Additional authors and addresses can be added with ``\and'',
% just like the second author.
% To save space, use either the email address or home page, not both
\and
Walterio W. Mayol-Cuevas\\
Bristol University\\
% First line of institution2 address\\
{\tt\small wmayol@cs.bris.ac.uk}
}

\maketitle
%\thispagestyle{empty}

%%%%%%%%% ABSTRACT
\begin{abstract}
In this paper, we address the problem of estimating the hand pose from the egocentric view when the hand is interacting with objects. Specifically, we propose a method to label a dataset Ego-Siam which contains the egocentric images pair-wisely. We also use the collected pairwise data to train our encoder-decoder style network which has been proven efficient in \cite{xiao2018simple}. This could bring extra training efficiency and testing accuracy. Our network is light weight and can be performed with over $30$ FPS with an outdated GPU. We demonstrate that our method outperforms Mueller et al. \cite{mueller2018ganerated} which is the state of the art work dealing with egocentric hand-object interaction problem on the GANerated \cite{mueller2017real} dataset. To show the ability to preserve the semantic information of our method, we also report the performance of grasp type classification on GUN-71 \cite{rogez2015understanding} dataset and outperforms the benchmark with only using predicted 3-d hand pose.

\end{abstract}

%%%%%%%%% BODY TEXT
\section{Introduction}
With the development of first person view headsets and action cameras, understanding the hand's activity is of great importance for human-computer interaction. Acquiring the hand pose could help people to explore the relationship between hand and environment further. In order to estimate 2d/3d hand pose and shape. People have made lots of efforts on both depth image based \cite{ge20173d,oikonomidis2011efficient,keskin2012hand,keskin2013real, sinha2016deephand, sharp2015accurate, taylor2016efficient,tang2014latent} and monocular based \cite{ge20193d,zimmermann2019freihand, hasson2019learning, hampali2020honnotate, tekin2019h+} methods. Most of these works achieve good performance on hand pose estimation precision. However, the real-time performance and hand-object interaction robustness are not considered as the primary factors in performance evaluation. As the work from Hasson et al. \cite{hasson2019learning}, they try to reconstruct 3d hand and object shape and put more focus on minimising the contact loss between hand and objects. Unsurprisingly, a better result is achieved with training on the synthetic dataset. For the real dataset, due to the uncertainty in occlusion and errors accumulated in data labelling, it is hard to get the real ground truth for training and evaluation.  
 \begin{figure}[t]
  \includegraphics[width=\linewidth]{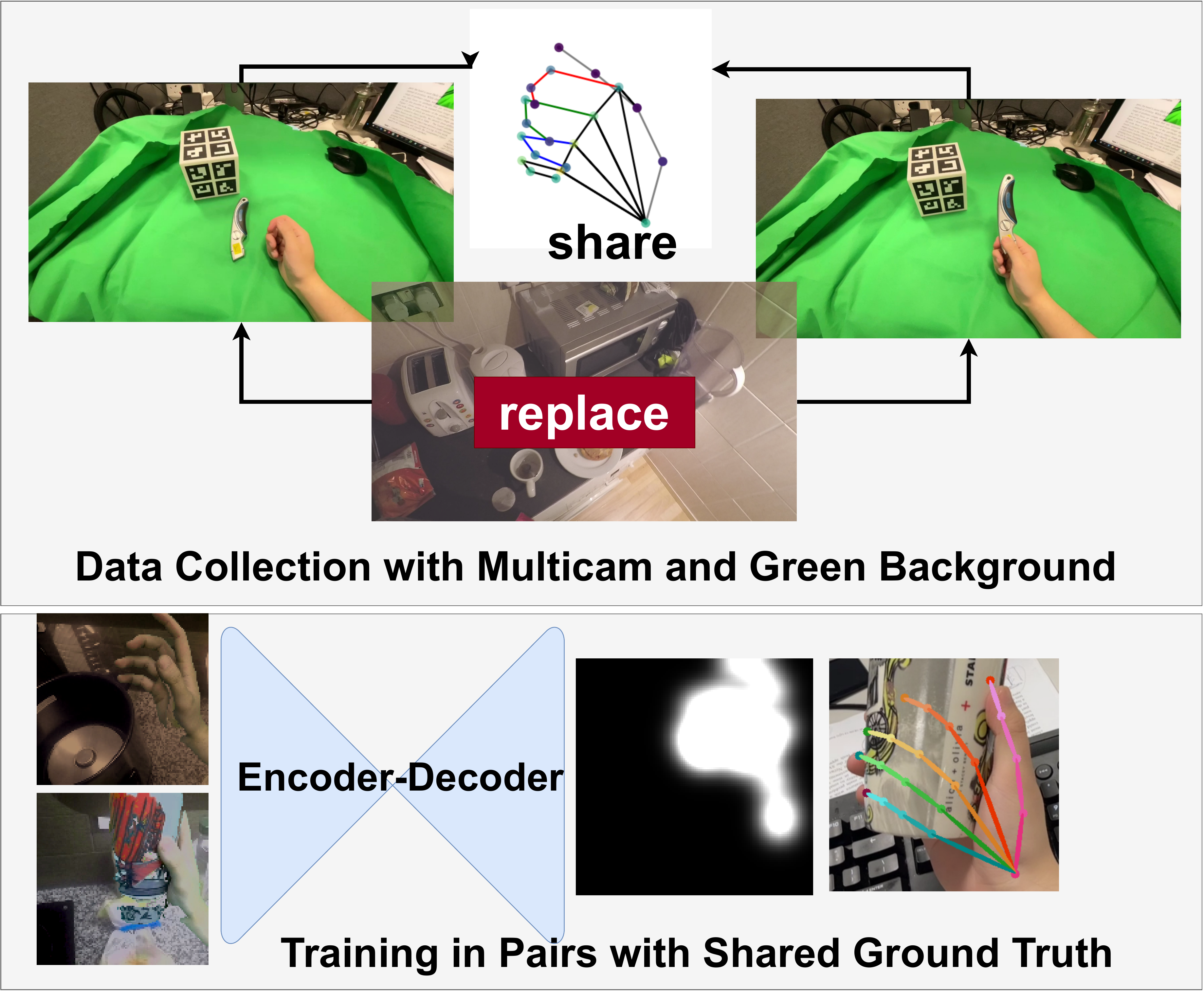}
  \caption{The overview of our work. We first use a multi-cam system and a replaceable green screen to collect the data in occlusion and non-occlusion pairs. The data is use to train our network pair-wisely.}
  \label{fig:overview}
\end{figure}
Particularly in the egocentric view, the camera mimics the eye of the camera wearer. The hand is of more importance on activity recognition and understanding. Moreover, the upgrading of wearable cameras in recent years makes the egocentric vision clearer and broader visual perception ability that can capture more details on hand-object interactions. However, the egocentric vision also brings many challenges to hand pose estimation. Like hand-hand, hand-object and self-occlusion could bring huge uncertainty on hand pose estimation. Furthermore, to our knowledge, besides the work of Garcia et al. \cite{garcia2018first}, there are no real dataset targeting handling hand-object interaction problems. Nevertheless, ego-vision provides a relatively fixed viewpoint on hands, reducing the complexity of modelling hand orientation.

Because our purpose for hand pose estimation is not for hand-virtual object contact reasoning or hand-object interaction reconstruction, achieving high accuracy on hand joints is not the primary concern. Instead, we argue that estimating pose for recognising and reproducing the semantic grasp type is more meaningful for human-centred activity understanding.

In this work, we mainly address the problem of estimating the hand pose under occlusion caused by hand-object interaction. We setup a highly configurable multi-cam system for hand data collection and collect our dataset in a novel pair-wise style. Our main contribution of this paper can be list as below:
\begin{itemize}
    \item Propose a highly configurable multi-cam system combining a novel pair-wise data collection pipeline.
    
    \item Annotating a new dataset Ego-Siam for egocentric hand pose estimation under object occlusion and cluttered background.
    
    \item Training a encoder-decoder based network for 2d/3d hand pose estimation with a novel pair training style.
    
    \item Evaluating the performance of 3d hand pose for grasp type classification.
    
\end{itemize}

\section{Related Work}
In this work, our purpose is to solve the occlusion problem by introducing a new dataset and training method with a simple network. We mainly review the current RGB based datasets and SOTA methods for hand pose estimation in the circumstances of object manipulation.

To solve the 2-d hand pose estimation problem in egocentric perspective, people annotated real data \cite{rogez2015understanding,mueller2017real,garcia2018first} and created synthetic data \cite{mueller2017real,mueller2018ganerated,lin2021two,malik2019simple}. All of them provide 3-d hand pose ground truth. However, the type and quality of annotations vary. From the summary of table \ref{tab:dataset}, we found the real data an egocentric perspective is very scarce. The UCI-EGO \cite{rogez2015understanding} is annotated by manual refinement. As far as we could find, it is no longer available online. FHAD provides massive labelled 3-d hand-object manipulation data. However, the RGB frames are irreparably damaged by the magnetic sensors attached on hand for data collection. And for EgoDexter \cite{mueller2017real}. The quantity and quality of data annotated are very limited. For each frame, only the visible fingertips are manually labelled. Because the annotator labelled from depth images, the dis-alignment between the depth and RGB also leads to inaccuracies. The synthetic datasets provide abundant poses, objects and backgrounds, and they still suffer from the unrealistic shapes, poses and skin textures. Providing high-quality hand joints annotations in an egocentric perspective still remains challenging.          

Hand pose estimation has been studied extensively in recent years. Traditionally, the methods can be categorised into model-based and appearance-based. With the development of CNN (Convolutional Neural Network), the boundary between the two kinds of approaches has been blurred. Zimmermann and Brox \cite{zimmermann2017learning} first predict 2-d hand pose and estimate 3-d pose by finding the most similar priors in 3-d space. Mueller et al. \cite{mueller2018ganerated,mueller2017real} use a 3-d skeleton to fit the detected 2-d hand joints. And recently, people use a trainable mesh convolutional decoder to decode the hand shape and pose \cite{kulon2020weakly, hasson2019learning,romero2018speeded}. In summary, we found recent works put more attention on data collection instead of models. This indicates the importance of training data. The difference in training data could bring a huge performance gap on different applications. And the result from Bin et al. \cite{xiao2018simple} that the simple network is able to have relatively good performance further weaken the role of the network in hand pose estimation. For the egocentric hand-object manipulation problem we are addressing in this paper, as far as we know, there is no real data for training purposes available. In this work, we focus on estimating 2-d hand landmarks only, which is still fundamental in 3-d hand pose estimation and data collection.

\begin{table}
\begin{center}
\begin{tabular}{|c|c|c|c|c|c|}
\hline
 Dataset & S/R  & HOI   & Frames \\
\hline\hline
UCI-EGO\cite{rogez2015understanding}          & Real & None   & 400         \\
SynthHands\cite{mueller2017real}       & Synth & both   & 63,530   \\
EgoDexter\cite{mueller2017real}       & Real & obj   & 1485     \\
GANerated Hands\cite{mueller2018ganerated}     & Synth & both   & 330k    \\
FHAD\cite{garcia2018first}            & Real & obj   & 100k     \\
SynHandEgo\cite{malik2019simple}      & Synth & None   & -         \\
Ego3DHands\cite{lin2021two}     & Synth & None   & 50k/5k      \\

\hline
\end{tabular}
\end{center}
\caption{The table shows the recent datasets with egocentric data for hand pose estimation. 'S/R': Synthetic or Real dataset. 'HOI': Contains hand-object interaction data.}
\label{tab:dataset}
\end{table}

\section{Method Overview}
We aim to predict the 3-d hand pose in an egocentric view, especially for the cases of hand-object interaction. To achieve this, we contribute the work in two steps: data collection and training (shown in figure \ref{fig:overview}). For data collection, we setup a multi-cam system that utilising multiple monocular cameras and 2-d keypoints detectors to acquiring 3-d hand pose annotation. This process is implemented with a green screen as background. These collected 3-d data are re-projected back to the egocentric view and used for refining the 2-d hand keypoints detector. The pipeline is first used iteratively in \cite{simon2017hand} and improved in \cite{zimmermann2019freihand, hampali2020honnotate}. After obtaining the optimised 2-d hand detectors, we use a novel method that can significantly reduce the uncertainties caused by object occlusion. The data is collected in a pair-wise style. A pair of data contains two images. They share the same ground truth. One of them poses the interaction grasp type with an object in hand. The other image has everything the same as the previous one except the object is removed from the hand. For training, we feed our encoder-decoder style network with paired data for 2-d heatmaps prediction. The 3-d hand pose is predicted by an MLP (Multi-Layer Perceptron) following the predicted heatmaps.
% With the paired data, we propose a Siamese training pipeline that enable the network to 'ignore' the occluded part of hand. We test our detector on EgoDexter \cite{mueller2017real} quantitatively. To facilitate preciser evaluation. We also collect a new test set which contains hand-object interactions with different occlusion level.
 
\begin{figure}
  \includegraphics[width=\linewidth]{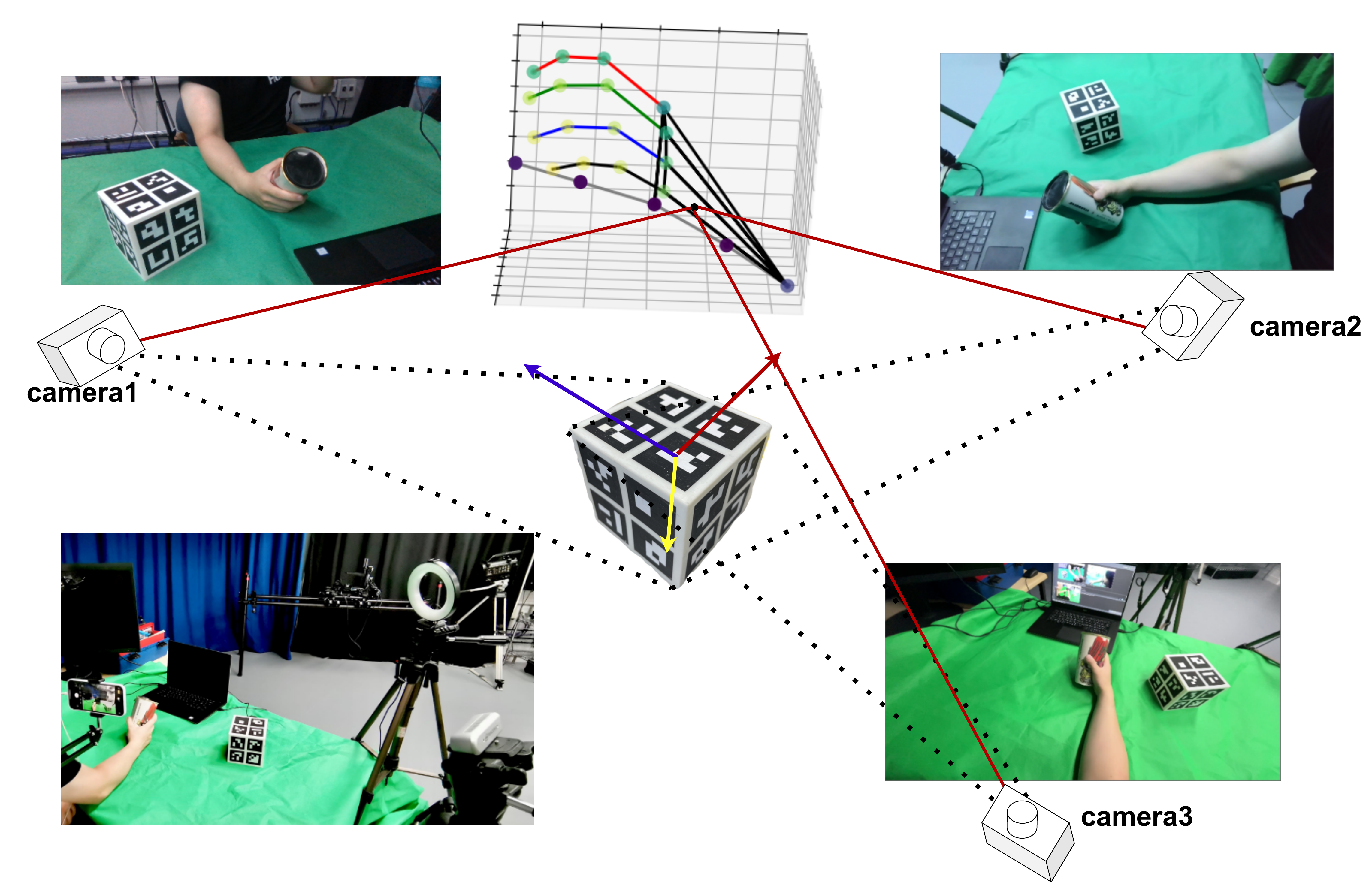}
  \caption{The Multi-cam system set-up for 3-d hand pose annotation. Each camera can obtain multiple 6-d positions (3 for location and 3 from orientation) from the detected markers on the cube. By fitting a skeleton to the observed 2-d data, we can find the optimum 3-d hand joint position. 
  }
  \label{fig:setup}
\end{figure}

\begin{figure}
  \includegraphics[width=\linewidth]{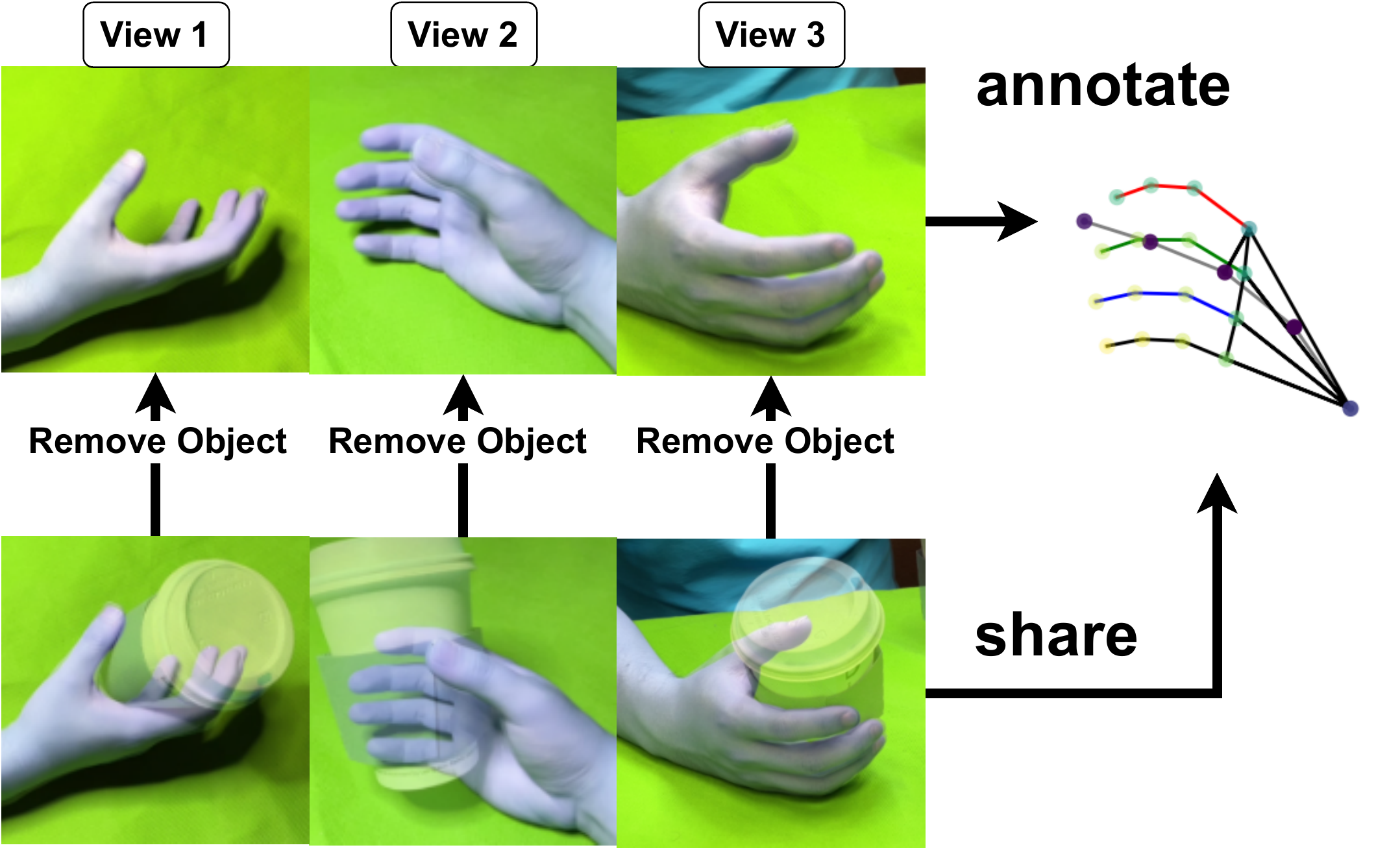}
  \caption{The figure shows the pipeline of data collection. The fist set of data is captured \textbf{with} object in hand. The two sets share the same ground truth which is annotated based on the second set (without object).}
  \label{fig:siam_data}
\end{figure}
\subsection{Multi-cam Setup}
We setup a highly configurable and extensible multi-cam system for data collection (Shown in figure \ref{fig:setup}). Our current setup contains $2$ normal HD cameras and an iPhone $12$ which has a wide-angle lens (used for capturing egocentric data). Cameras can be set from any viewpoint without frequent calibration of extrinsic parameters. Instead of fixing cameras from beginning to end, we leverage the ArUco code \cite{romero2018speeded} (a library for fast visual localization) for online calibration, which returns the camera 6-d pose relative to marker centre during capturing. 
\begin{figure}
  \includegraphics[width=\linewidth]{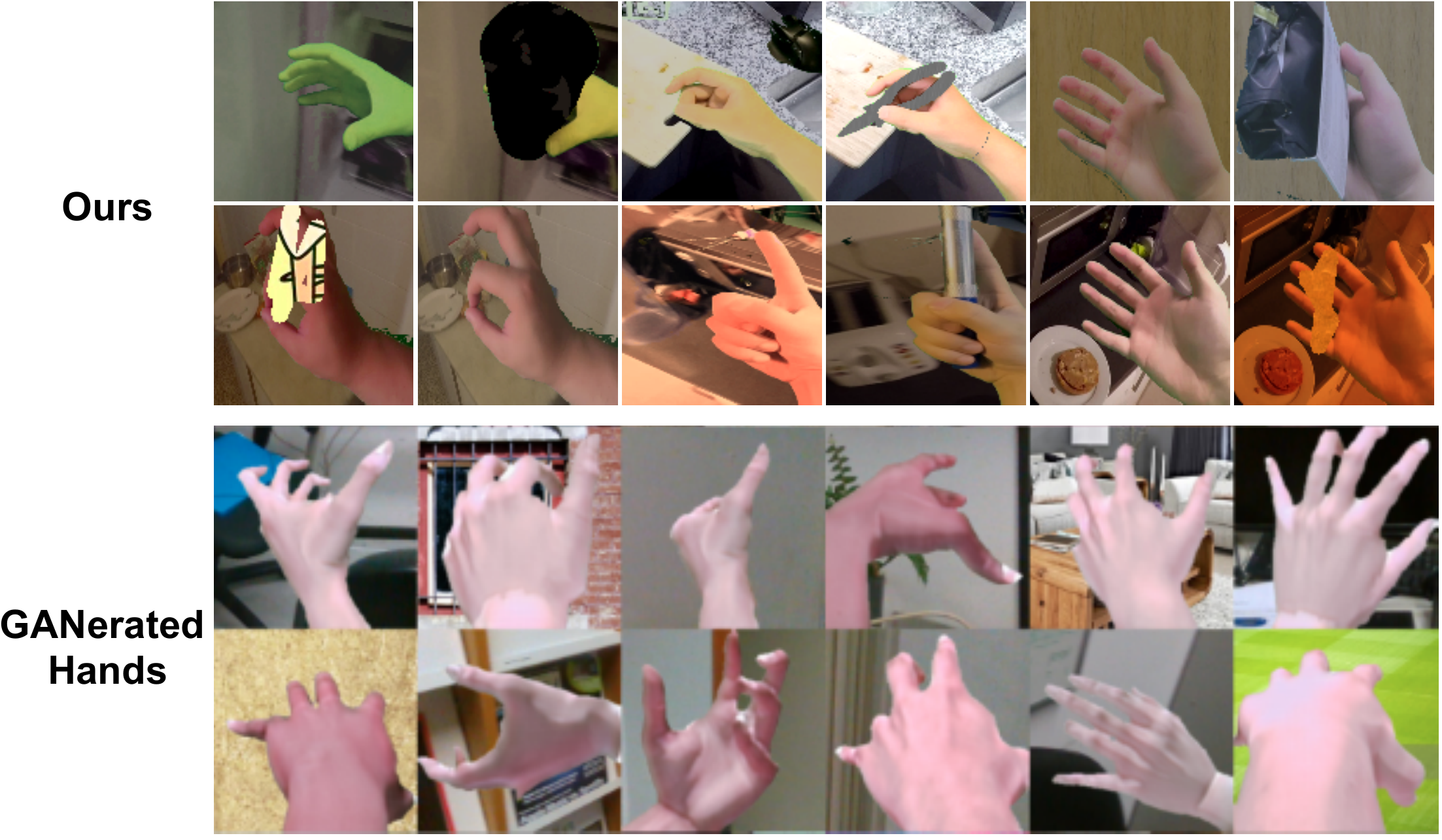}
  \caption{The figure shows the visual comparison between our augmented data and the GANerated Hands \cite{mueller2018ganerated}. Our dataset is more realistic and involves more hand-object interactions.}
  \label{fig:siam_data}
\end{figure}

However, relying on the localization result from a single ArUco marker may bring occlusion to particular camera views and errors. In order to increase the robustness of ArUco code recognition from different views, a 3-d printed ArUco cube is used (the cubic object shown in figure \ref{fig:setup}). The cube has six sides, and four different ArUco markers adhere to each side. Every marker's position is precisely defined in advance, and their centres are offset to one corner of the cube (the axis on the cube shown in figure \ref{fig:setup}). This means every visible marker could return the camera 6-d position relative to the corner coordinate system of the cube. To deal with multiple 6-d poses returned by cube, we use RANSAC to find the optimum.

Before data collection, the ArUco cube is placed in a position where is visible for all the cameras. The 6-d poses of the cameras relative to the cube coordinate system are obtained as extrinsic parameters. During the data collection, the camera \textbf{\textit{1}} wearer also needs to keep the ArUco cube appeared in the scene, and the 6-d extrinsic parameters will be determined frame-wisely.

To preserve hand grasps' RGB appearance, instead of using a marker or invasive tracker on hand \cite{garcia2018first}, we detect 2-d keypoints on the images observed from different cameras for 3D data generation. To achieve a fast and automated labelling process, a 2-d keypoints detector with the spirit of the work \cite{xiao2018simple} is used for initialization. The detector is pre-trained on the GANerated \cite{mueller2018ganerated} dataset (only in the data collection stage). Like the work \cite{simon2017hand}, we apply a bootstrapping procedure with an external source of physical hand model supervision and iteratively optimize the detector. The hand model is designed with froward kinematics which can be parameterized by $\Omega=(\theta, \gamma, \phi)$. $\theta\in \mathbb{R}^{20}$ controls the movement of each finger. $\gamma\in \mathbb{R}^{5}$ controls the length of each finger. And $\phi\in \mathbb{R}^{6}$ controls the global position and orientation of the hand. $\gamma\in \mathbb{R}^{5}$ are constants for a certain person and need to be measured in advance with fingers spread. Our task is to find an optimum configuration of $\Omega$, which has the minimum discrepancy between our hand model and observed 2-d keypoints from each view. Furthermore, a hand mask is used to reduce the importance of occluded joints in each view. The optimization problem can be re-expressed as minimizing the loss function: 
$$Loss = Loss_{2d}+Loss_{model}   {(1)}$$
$Loss_{2d}$ is determined by summing the distance between detected 2-d joints position $j_{k}^{i}\in \mathbb{R}^{2}$ and 3-d hand model joints $J_{k}\in \mathbb{R}^{3}$ projection across all views $i=\{1, 2, 3\}$:
$$Loss_{2d} =v_{k}^{i}\omega_{k}^{i}\sum_{k}\sum_{i}||j_{k}^{i}-Proj^{i}(J_{k})||_2{(2)}$$where $\omega$ is the confidence from 2-d joint detector and $k$ is the number of joints. And $v_{k}^{i}$ is the weights to penalize the importance of joints occluded. If the joint $j_{k}^{i}$ is within the mask 
detected in view $i$, $v=1$, otherwise, $v=0.5$. The last term $Loss_{model}$ is used to physically constrain the hand model for generating meaningful poses:
$$Loss_{model} =\sum_{j}( \beta){(4)}$$$\beta =0$, if $x=\theta_{j} \in \{\theta_{j}^{lower\_limit}, \theta_{j}^{upper\_limit}\} $. \\
$\beta =constant$, if $x=\theta \not\in \{\theta_{lim}^{lower}, \theta_{limit}^{upper}\} $\\
\\
We use \textit{Levenberg–Marquardt} algorithm \cite{more1978levenberg} to optimize the loss function which has been proven efficient in many works \cite{taylor2016efficient} \cite{mueller2018ganerated} \cite{ge20173d}. In the beginning, the 2-d detection of keypoints can be inaccurate. The frames with an unreasonable error are identified and relabelled manually. After the first labelling round, we labelled 3-d data and corresponding 2-d projection on each view. The 2-d projection is used to fine-tune our 2-d detector. Similar to the work \cite{simon2017hand}, this process is executed several times until the error becomes acceptable.

\subsection{Siamese Data Collection}
For a 3-cam system, eliminating the inaccuracy caused by the hand-object interaction is still challenging. Especially from the egocentric perspective, like the grasp type 'large diameter' (grasping a large object with the whole palm and fingers), the majority of the hand can be occluded from an egocentric perspective. This usually leads to catastrophic failure on 2-d keypoints detection. We guess this could be the reason why the hand pose estimation for egocentric hand-object interaction remains unsolved even a lot of multi-cam data collection systems have been proposed in the community.

To tackle this problem, we propose a simple and effective solution. We capture the data in pairs. Each pair has two sets of frames. The whole process is shown in figure \ref{fig:siam_data}. With the participator performing the hand-object interaction on one hand, we capture the frames from multi-cam as the first set of data of a pair. And then, keeping the hand as still as possible, the second set of the data pair is obtained after the object is removed. The two sets of images compose a pair of data that shares the same ground truth from the multi-cam annotation on the second set of frames (the object is removed in the second set).
To increase accuracy, we overlay the semi-transparent image from the previous set on the screen to make hand alignment between two sets easier. In data annotation, we found that estimating the hand pose with occlusion is almost 'guessing' the position of the occluded joints with the visible part. Particularly from the egocentric perspective, the occluded parts have high variability while keeping the visible appearance same. To this end, in data annotation, we try to perform the hand-object interaction with canonical grasp types proposed in \cite{feix2015grasp} to reduce the uncertainty of posterior distribution.

The data collection process requires the camera and hand to keep still, instead of mounting the camera on the head or chest. We use a retractable link to mimic the egocentric vision. Benefit from the online calibration of our system. The position of the egocentric camera can be frequently changed during the data collection. We also use a fill-in light to simulate different light conditions.

\subsection{Dataset and Data Augmentation}
\label{sec:data_aug}
We collect $2k$ pairs of hand-object interaction data with a single right hand from a male. To gain more diversity on hand shapes and colours, another $2k$ frames without hand-object interaction are collected with $2$ male hands and $2$ female hands. There are $6k$ frames in total as training data. For testing, we collect another $500$ frames with hands performing different grasp types (the hand did not appear in the training set). With the green screen, the data augmentation can introduce more variability in the background. We identify the green colour in HSV colour space and replace it with real egocentric scenes, including the frames from EPIC-KITCHENS \cite{damen2018scaling}, GTEA \cite{fathi2011learning} and the frames we collected from our offices and kitchens. Some backgrounds in skin colour are also added. Besides, we manually label the object masks for several hundreds of frames and replace them to increase the variability on object appearance.

\begin{figure}
  \includegraphics[width=\linewidth]{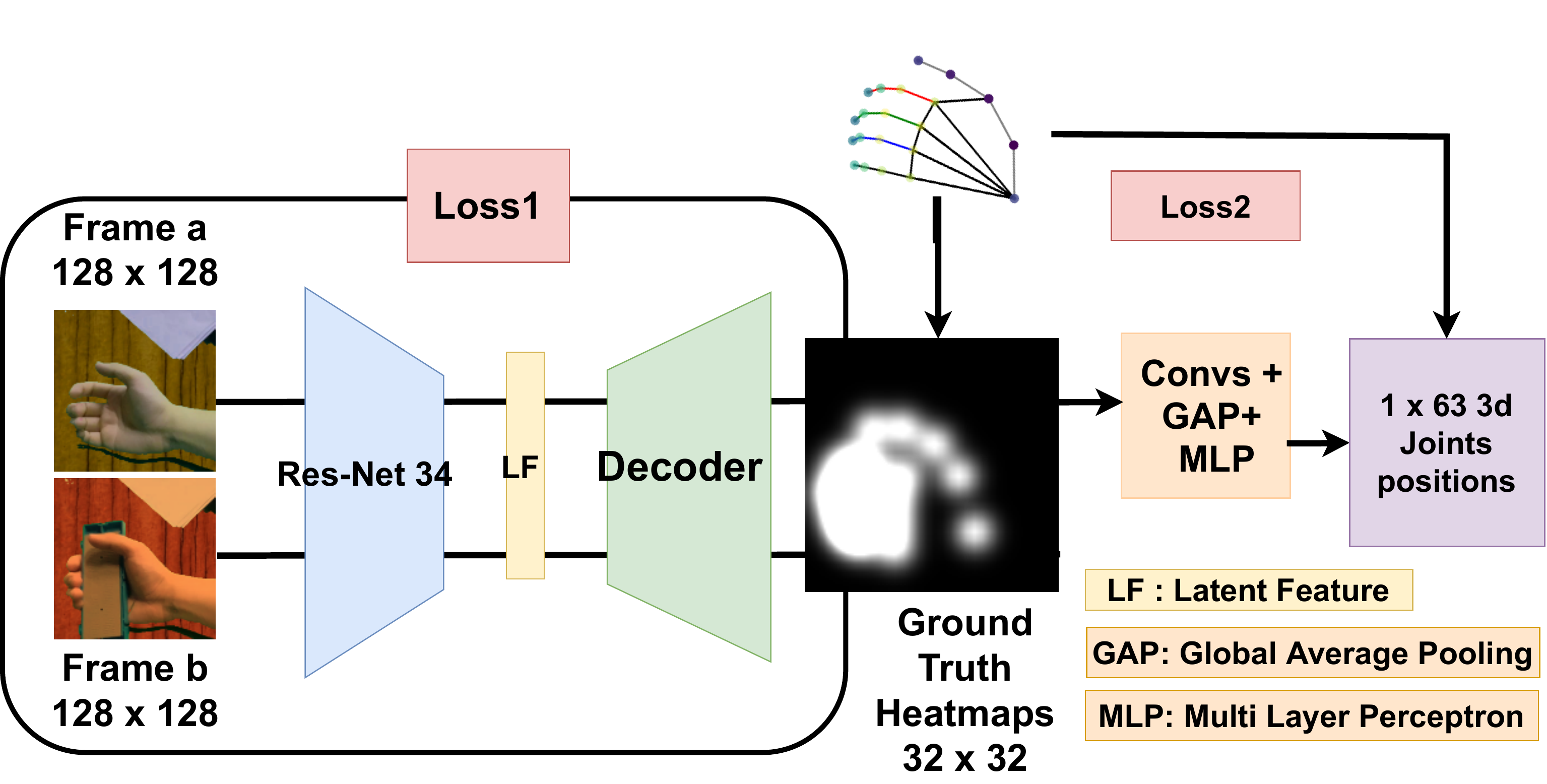}
  \caption{The overview of our network. We use a 34 layers Res-Net as encoder and used multiple up-sampling layers as decoder. The data is fed in pairs. To predict the 3-d hand pose, we add a global average pooling layer and a multi-layer perceptron after the predicted heatmaps.}
  \label{fig:siam_training}
\end{figure}

Another important augmentation is putting artificial occlusion on images. We randomly add line linkages between joints and circles with random sizes to simulate object occlusion. Other standard operations like random contrast, random brightness and random warp are also applied. All the augmentations and texture replacements run with the process of training. 
\begin{figure*}[t]
  \includegraphics[width=\linewidth]{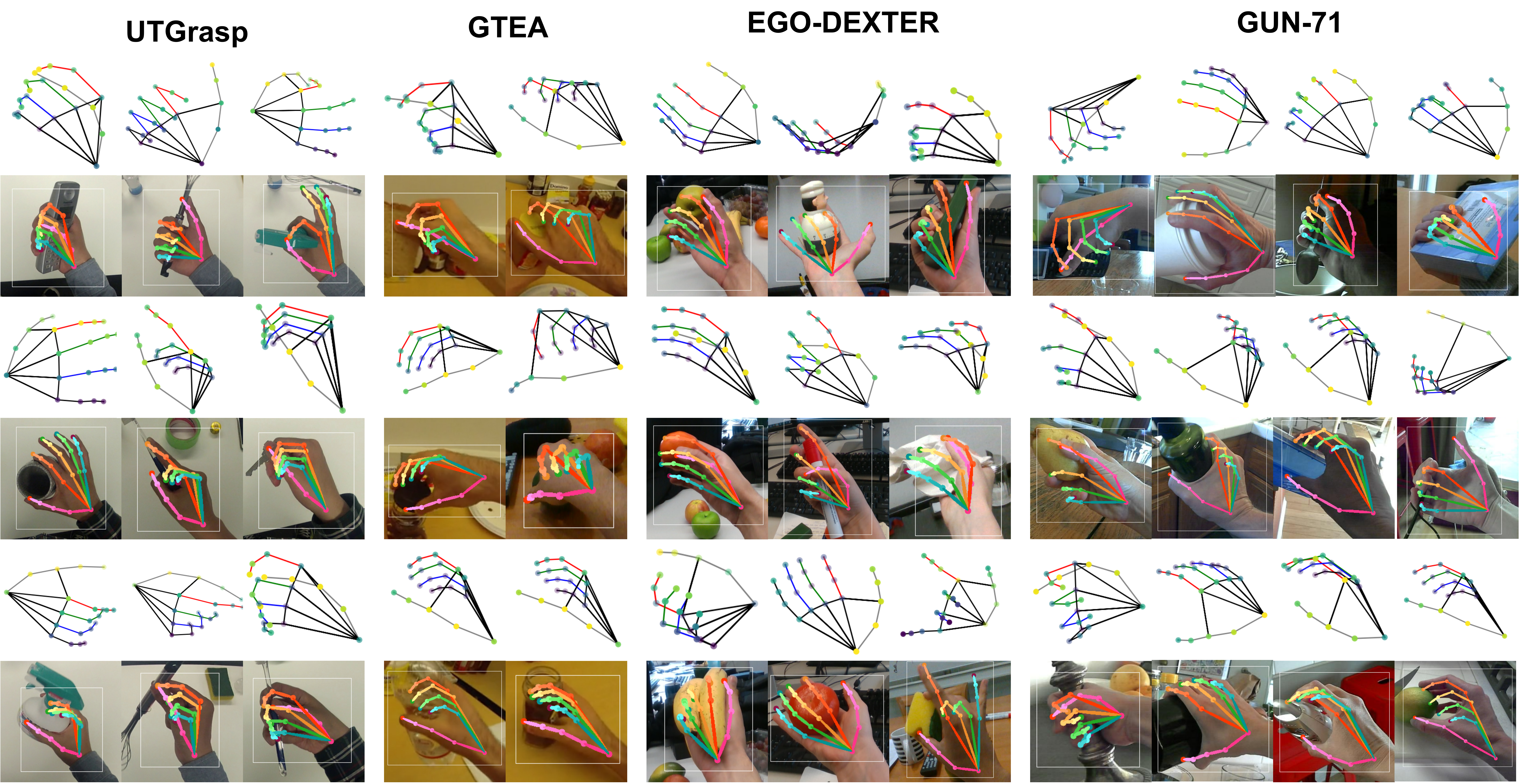}
  \caption{The visual results on UTgrasp \cite{cai2015scalable}, GTEA \cite{fathi2011learning}, EgoDexter \cite{mueller2017real} and GUN-71 \cite{rogez2015understanding} of our 2d/3d hand pose estimator. Our method can work with different hands, light conditions and objects.}
  \label{fig:visual_rst}
\end{figure*}
\subsection{Training in Pairs}
It is harder to make the network learn all the configurations with limited data. Particularly, for learning to predict hand pose under occlusion, it is impractical to consider every hand-object combination in training data. Besides making good variability with the data augmentation we detailed in section \ref{sec:data_aug}, we also want the network to associate the full hand pose when the hand is only partially visible. The work from zhou et al. \cite{zhou2020occlusion} provides a solution. They use a Siamese network to reconstruct the feature occluded by objects. However, the reconstruction module does not help with gaining a significant improvement on accuracy. We believe that instead of keeping the features on occluded parts, ignoring them is more practical and efficient. After several experiments, we found that training our pair-wise data in pairs (back-propagate the loss from the pair of data and train the network with standard configurations) can remarkably improve the training efficiency. Experiments can be found in section \ref{sec:ablation}.

% We come up with a novel Siamese training method to training our network. The Siamese net (also called as twin net) is widely used in object recognition \cite{koch2015siamese} and object tracking \cite{bertinetto2016fully}. The idea is using the same weights with different inputs. 

As shown in figure \ref{fig:siam_training}. We input the two images from one pair of data. One image has an object/artificial occlusion on hand. One has nothing on hand. The pair of data share the same ground truth. We want the encoder-decoder structure to 'learn' the common and differences by back-propagating the loss from both images in each training step. Due to the characteristics of the egocentric view, the 3-d pose ambiguity brought by the 2-d prediction is greatly constrained. We predict a 3-d hand pose using a very small network (a multi-layer perceptron) that only takes 2-d heatmaps as input. The experiments and results in \ref{sec:ablation} show that training in pairs does improve the training efficiency and network performance.

\begin{figure}
  \includegraphics[width=\linewidth]{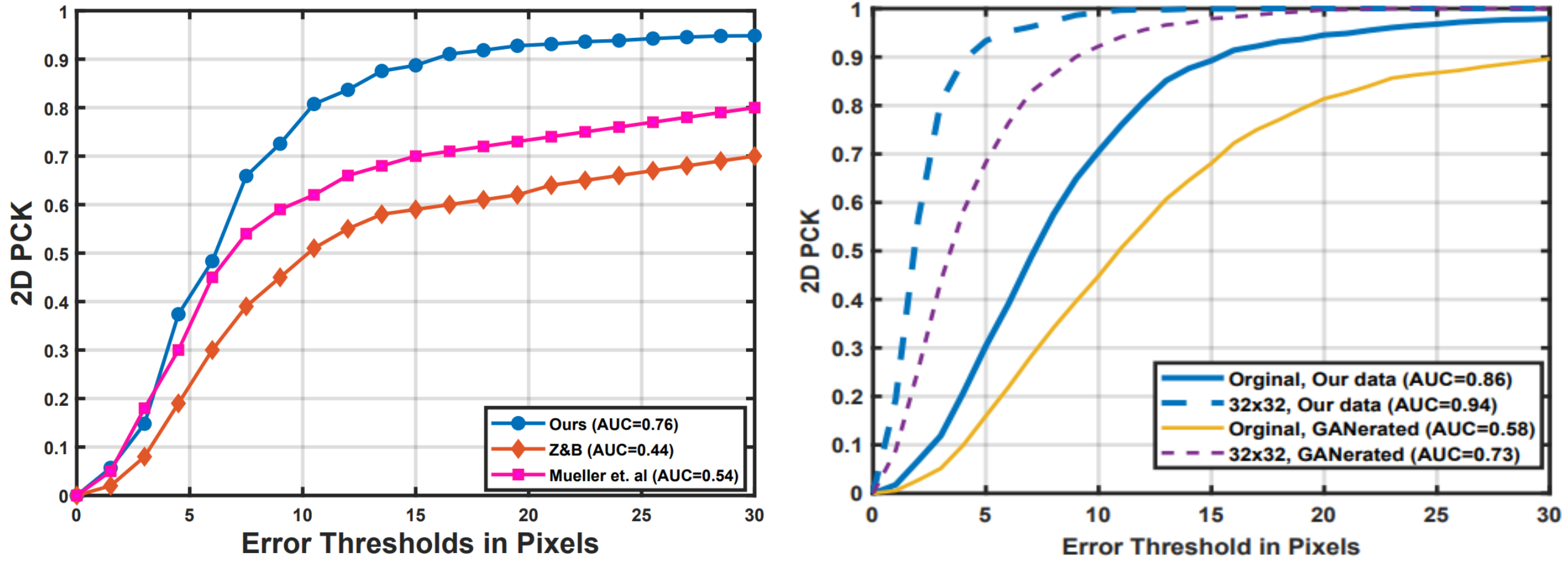}
  \caption{The quantitative results on EgoDexter \cite{mueller2018ganerated} and our Ego-Siam test set.}
  \label{fig:auc_curve}
\end{figure}

\section{Experiments and Results}
In the experiments section, we evaluate our approach and dataset quantitatively and visually. Same as the evaluation protocol used in pose estimation works, The Percentage of Correct Keypoints (PCK) score is used as our evaluation metric. It measures whether the given keypoint falls within a pre-defined range around the ground truth. 
Because our 3d hand pose is estimated according to the predicted 2-d heatmaps only, we mainly report the PCK plot of 2d results. And due to the limited availability of real hand pose datasets in an egocentric perspective. We can only conduct the quantitative evaluation on EgoDexter \cite{mueller2017real} and our new collected Ego-Siam test set. Besides, to research how much semantic information of hand can be preserved with the predicted hand pose, we use our trained model to detect the hand pose in the Grasp Understanding Dataset (GUN-71) \cite{rogez2015understanding} and training $17$-class and $33$-class classifiers for evaluating the performance on grasp type recognition.

\subsection{Results on Pose Estimation}
The figure \ref{fig:auc_curve} shows the AUC (area under the curve) values on EgoDexter\cite{mueller2017real} and our Ego-Siam data with different setups. Specifically, we compare the 2-d PCK on EgoDexter between Z\&B \cite{zimmermann2017learning}, Mueller et al. \cite{mueller2018ganerated} and ours. Our method greatly outperforms the fingertip detection on EgoDexter dataset. To further evaluate the performance of full hand pose estimation, we also compare the results on our Ego-Siam test set with different training data. The AUC value of training on Ego-Siam outperforms training on GANerated \cite{mueller2018ganerated} on both original image resolution and $32\times32$ (the resolution of output heatmap) resolution. As for 3-d results, although our 3-d results are extracted from the 2-d heatmaps only, we still achieve average error of $\textbf{36.02}$ $\textbf{mm}$ on EgoDexter\cite{mueller2017real} which is a comparative result with their benchmark $\textbf{32.6}$ $\textbf{mm}$. Furthermore, the amount of data we used for training (6k) is much less than Mueller et al. \cite{mueller2017real} (63k) and \cite{mueller2018ganerated} (330k). The results illustrate that our dataset is refined and can be generalised to different datasets. We show the visual results in figure \ref{fig:visual_rst}. It includes the results on UTgrasp \cite{cai2015scalable}, GTEA \cite{fathi2011learning}, EgoDexter \cite{mueller2017real} and GUN-71 \cite{rogez2015understanding}. It further proves that our method (training with Ego-Siam and our network only) can work across different light conditions, hands and objects from different datasets.

\subsection{Does Training in Pairs helps?}
\label{sec:ablation}
\begin{figure}
  \includegraphics[width=\linewidth]{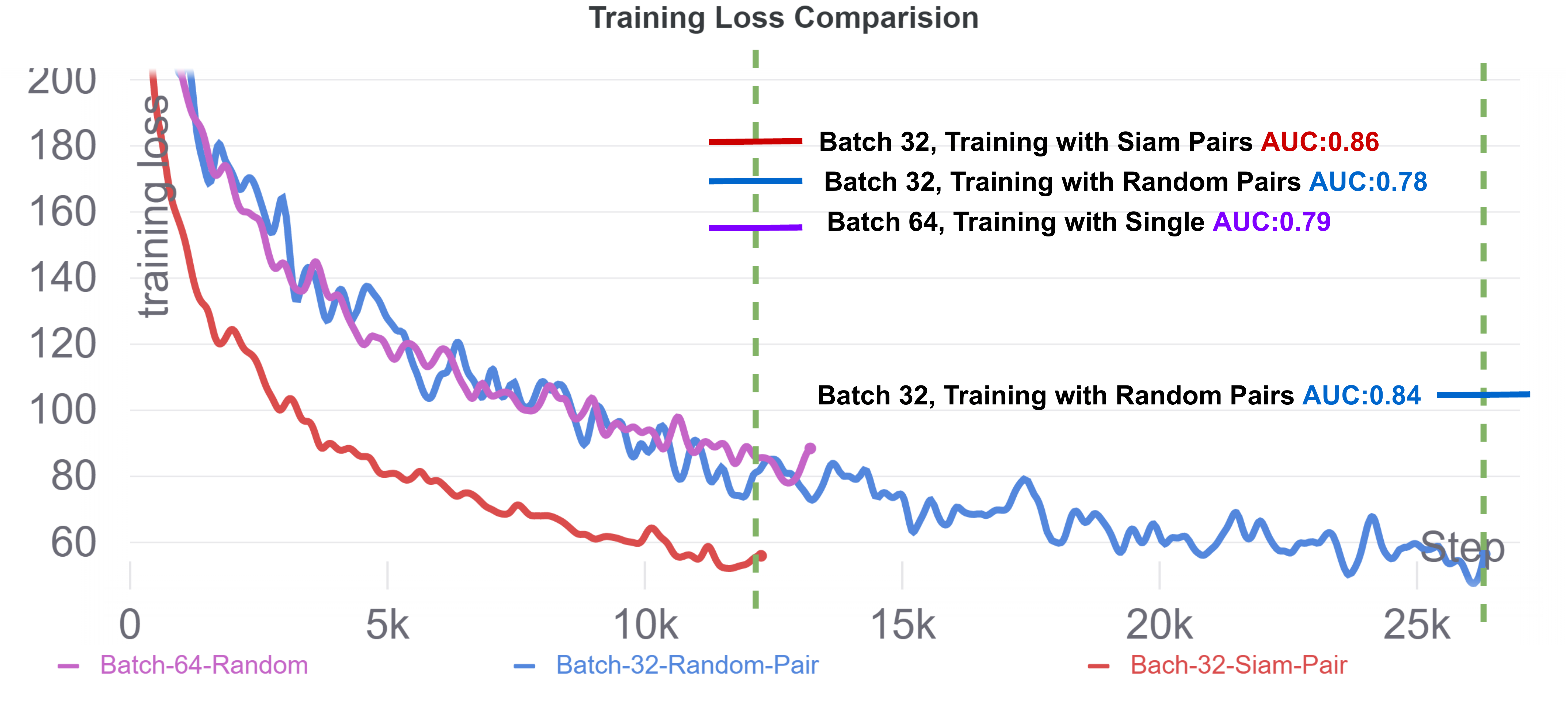}
  \caption{The training loss and corresponding testing results of the three experiment setups..}
  \label{fig:pair_training_loss}
\end{figure}

In our training setup, we train our network with a batch size of $32$. We have two forward passes for training data with and without object (including artificial ones) occlusion for each step, respectively. The loss for gradient descent comes from the sum losses of the two passes. To validate the effectiveness of our pair-wise training on Siamese data, we setup another two experiments. One follows the same training pattern. Differently, we randomly select two images as the pair. Another setup just use $64$ as batch size and training the network in the usual way. The figure \ref{fig:pair_training_loss} shows the change of training loss with time. We stop training the '32-batch Siam Pair' (red curve in the figure) at a step around $12k$ where we think the results on the testing set is acceptable. However, it takes the '32-batch Random Pair' nearly $27k$ steps to reach a similar performance. For 'Batch 64 Single', it has a similar performance with the '32-batch Random Pair'. The results show that training our Ego-Siam data does improve the training efficiency and accuracy.     

 With the Ego-Siam data, we believe there are more possibilities that can be explored. Like encoding the light conditions or background with an external model and training an encoder to isolate the non-joint information. We left these as future works.

\begin{figure*}[t]
  \includegraphics[width=0.9\linewidth]{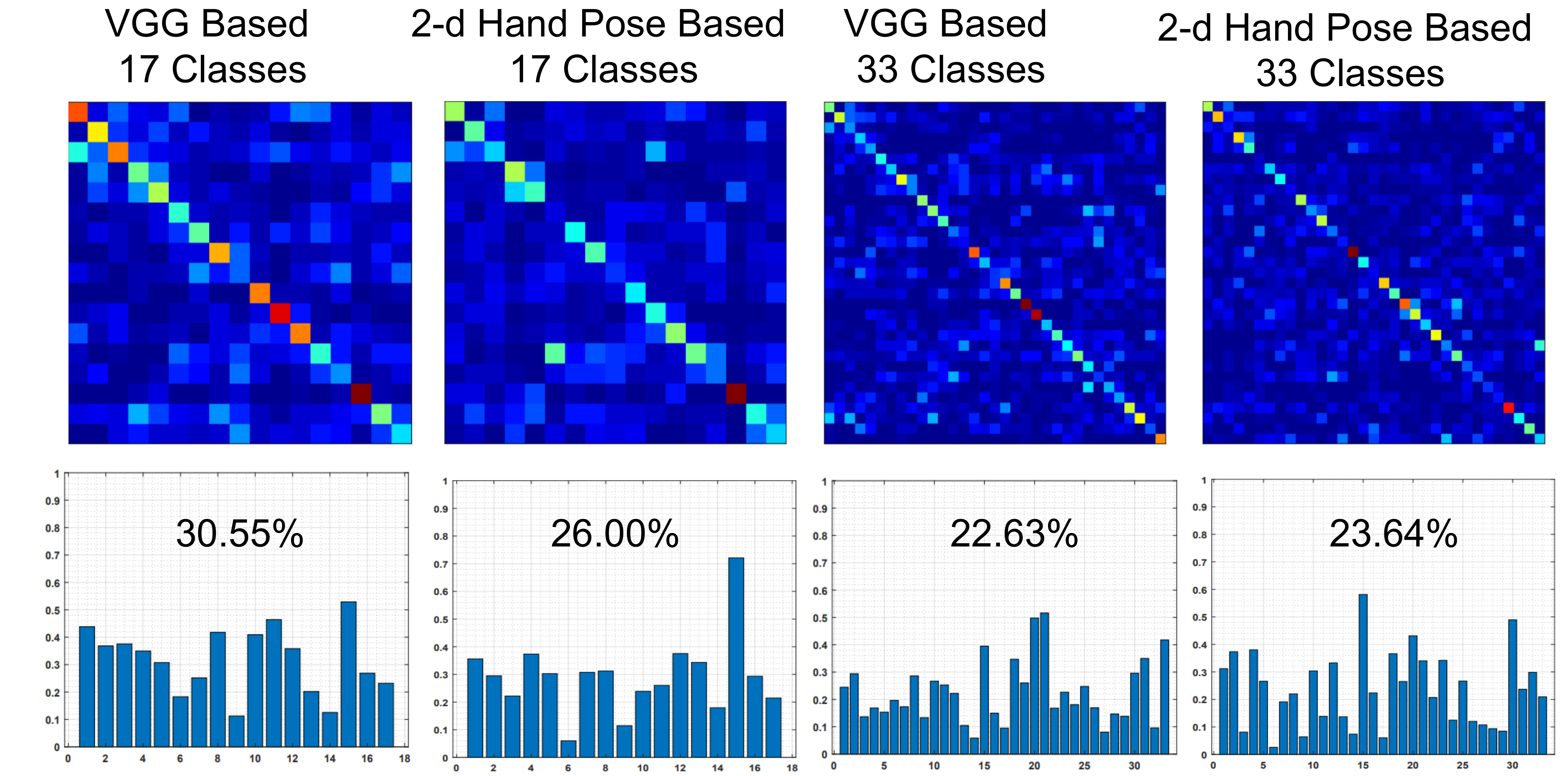}
  \caption{The confusion matrices and corresponding individual classification results.}
  \label{fig:Grasp_rst}
\end{figure*}
\subsection{Results on Grasp Recognition}
In most cases, the occluded fingers have enormous uncertainties. Evaluating the performance with a certain ground truth could be tendentious. We argue that a good pose estimation can preserve enough semantic information that is useful. The hand grasp type is a good representation that carries enough semantic information for evaluation.

We choose the Grasp Understanding Dataset (GUN-71) \cite{rogez2015understanding} to implement the experiments. There are $71$ kinds of grasp type in GUN-71. However, from the classes, $34-71$, the grasp type definitions greatly rely on the in-hand object instead of the way the hand handle it. Thus, we choose the 17-class taxonomy concluded by Cukosky et al. \cite{cutkosky1989grasp} and 33-class taxonomy from Feix et al. \cite{feix2009comprehensive} for evaluation. To training the classifier, we use a Faster-RCNN \cite{ren2015faster} based detector trained with 100-DOH dataset \cite{Shan20} to crop all the right hands in the dataset. We predict a $1\times63$ 3-d hand pose for each crop and feed it into a 3-layer MLP (multi-layer perceptron) as training data.     

The results is shown in table \ref{tab:grasp_rst}. We add a set of an experiment using a pre-trained VGG based classifier as a benchmark. The VGG based net takes the $128\times128$ hand crop as input and predicts the crop's grasp class. The results show that both our 3-layer MLP (multi-layer perceptron) and the VGG based image classifier outperforms the Rogez et al. \cite{rogez2015understanding}. For $33$-class classification, our method achieves the highest accuracy. Interestingly, in Rogez's work \cite{rogez2015understanding}, they also report the results of using VGG based network. We hypothesise the difference could come from different network configurations and pre-training operations. 

\begin{figure}
  \includegraphics[width=\linewidth]{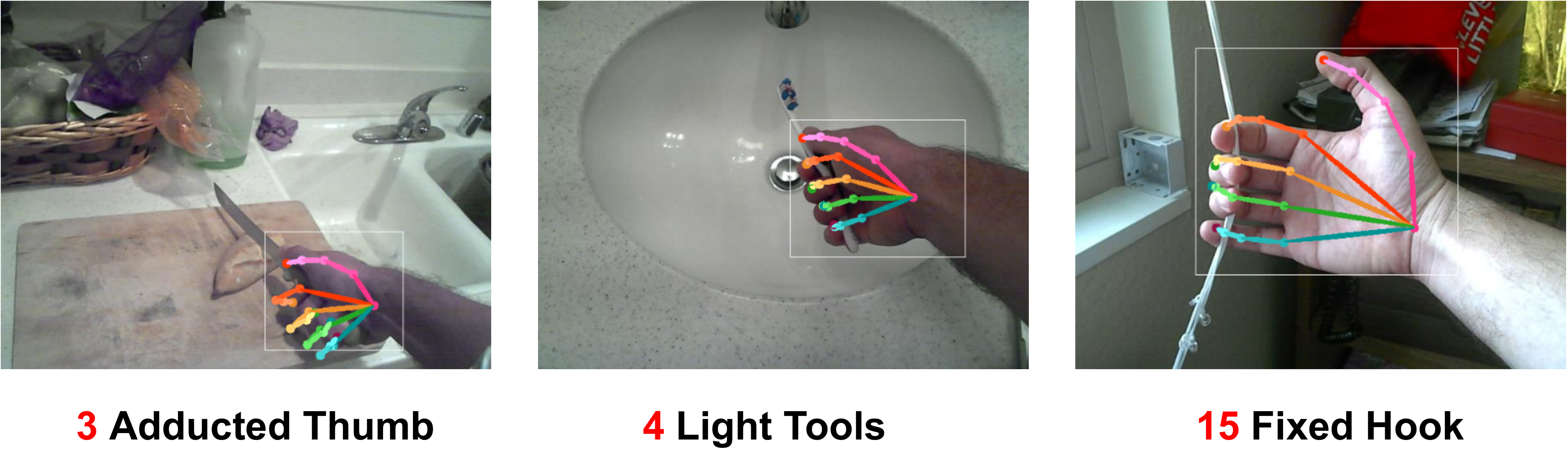}
  \caption{The class $3$ 'Adducted Thumb' and the class $4$ Light Tools can't be distinguished visually from egocentric view. The class $15$ 'Fixed Hook' is relatively easy to identify.}
  \label{fig:similarity_compare}
\end{figure}

The confusion matrices and their corresponding class-wise results are shown in figure \ref{fig:Grasp_rst}. Besides the failure on hand detection and hand pose detection, another important reason which causing the inaccuracy is the high visual similarity between different grasp types. By observing the confusion matrix, we pick the class $3$ 'Adducted Thumb' and 'Light Tool' as examples (Shown in figure \ref{fig:similarity_compare}). The two grasp types can not be even distinguished visually from the image. While for the class $15$ 'Fixed Hook', it is unique and easy to identify.

Overall, with only taking a $1\times63$ hand pose ($21$ 3-d joints) as input, our grasp type based classifier achieves competitive performance comparing with VGG based image classifier, which takes the image as input and having much more parameters. This illustrates that our hand pose estimator has a great ability to preserve the information on grasp types. By researching the grasp types that confuse classifiers, we have a chance to simplify the grasp taxonomy for egocentric perspectives.

\begin{table}
\centering
\caption{The averaged classification accuracy on $17$ and $33$ classes with different methods.}
\begin{tabular}{ |l|c|c|c| } 

\hline
     \#classes & Rogez et al.\cite{rogez2015understanding} & VGG based & Ours \\
\hline
% \multirow
     $17$\cite{cutkosky1989grasp} & 20.50\% & \textbf{30.55\%}& 26.00\%\\ 
    $33$\cite{feix2009comprehensive} & 20.53\% & 22.63\% &\textbf{23.64\%}\\ 

\hline
\end{tabular}
\label{tab:grasp_rst}
\end{table}

\subsection{The Real-time Performance}

We train and run inference on a desktop with Nvida Quadro M2000 GPU (4GB). The full hand pose estimation is performed in real-time with 27-35 FPS.

\section{Conclusion}
We present work that concerns the challenging problem of estimating egocentric hand pose under grasping occlusion. We propose a novel data collection pipeline and introduce a new hand pose dataset Ego-Siam contains $6000+500$ frames. To prove the efficiency of Ego-Siam dataset. We train an encoder-decoder style model with pair-wise data, which could make training easier. On EgoDexter \cite{mueller2017real} dataset, we achieve higher AUC value of $0.76$ on 2-d hand pose and comparative mean average error of $36.02$ $mm$ on 3-d hand pose comparing with the previous works. Through the evaluation of grasp type recognition based on GUN-71 \cite{rogez2015understanding} dataset, the ability to preserve semantic information of our hand pose estimator is proved.

{\small
\bibliographystyle{ieee_fullname}
\bibliography{egbib}
}

\end{document}